\documentclass[11pt,a4paper]{article}
\usepackage[margin=24mm,headheight=14pt,headsep=10mm,footskip=12mm]{geometry}
\usepackage[T1]{fontenc}
\usepackage{amsmath,amssymb,graphicx,booktabs,array}
\usepackage{cite}
\makeatletter
\providecommand\bstctlcite[1]{\@bsphack
  \@for\@citeb:=#1\do{%
    \edef\@citeb{\expandafter\@firstofone\@citeb}%
    \if@filesw\immediate\write\@auxout{\string\citation{\@citeb}}\fi}%
  \@esphack}
\makeatother
\usepackage{xcolor}
\usepackage{fancyhdr}
\usepackage{caption}
\usepackage{microtype}
\usepackage[hidelinks]{hyperref}
\hypersetup{pdfauthor={Zixuan Wang, Wei Zhang, Chenlei Zhu, Naiyu Chen, Xihua Wang, Jie Chen},pdftitle={CHASE: Curiosity-Guided Skill Acquisition for Embodied Task Execution},pdfsubject={Single-column technical report},pdfkeywords={robotics,curiosity,skill acquisition}}
\definecolor{chasenavy}{HTML}{233A50}
\definecolor{chaseblue}{HTML}{3A8298}
\definecolor{chasepaleblue}{HTML}{EDF4F7}
\definecolor{chaseline}{HTML}{C5D6DE}
\newcommand{\method}{CHASE}

\makeatletter
\renewcommand\section{\@startsection{section}{1}{\z@}%
  {-3.2ex plus -1ex minus -.2ex}{1.4ex plus .2ex}%
  {\normalfont\large\bfseries\color{chasenavy}}}
\renewcommand\subsection{\@startsection{subsection}{2}{\z@}%
  {-2.7ex plus -1ex minus -.2ex}{1ex plus .2ex}%
  {\normalfont\normalsize\bfseries\color{chasenavy}}}
\renewcommand\subsubsection{\@startsection{subsubsection}{3}{\z@}%
  {-2.2ex plus -1ex minus -.2ex}{.8ex plus .2ex}%
  {\normalfont\normalsize\bfseries\color{chasenavy}}}
\makeatother

\fancypagestyle{report}{%
  \fancyhf{}%
  \fancyhead[L]{\color{chasenavy}\footnotesize\sffamily\bfseries CHASE\enspace\color{chaseline}|\enspace\color{chasenavy}\mdseries TECHNICAL REPORT}%
  \fancyhead[R]{\color{chasenavy}\footnotesize\sffamily NOVEMBER 2026}%
  \fancyfoot[C]{\color{chasenavy}\footnotesize\thepage}%
}
\fancypagestyle{reportfirst}{\fancyhf{}\fancyfoot[C]{\color{chasenavy}\footnotesize\thepage}}
\begin{document}
\bstctlcite{IEEEexample:BSTcontrol}
\thispagestyle{reportfirst}
\noindent\begin{minipage}[c]{.45\textwidth}
\includegraphics[trim=143.19bp 217.47bp 143.19bp 217.47bp,clip,width=38mm]{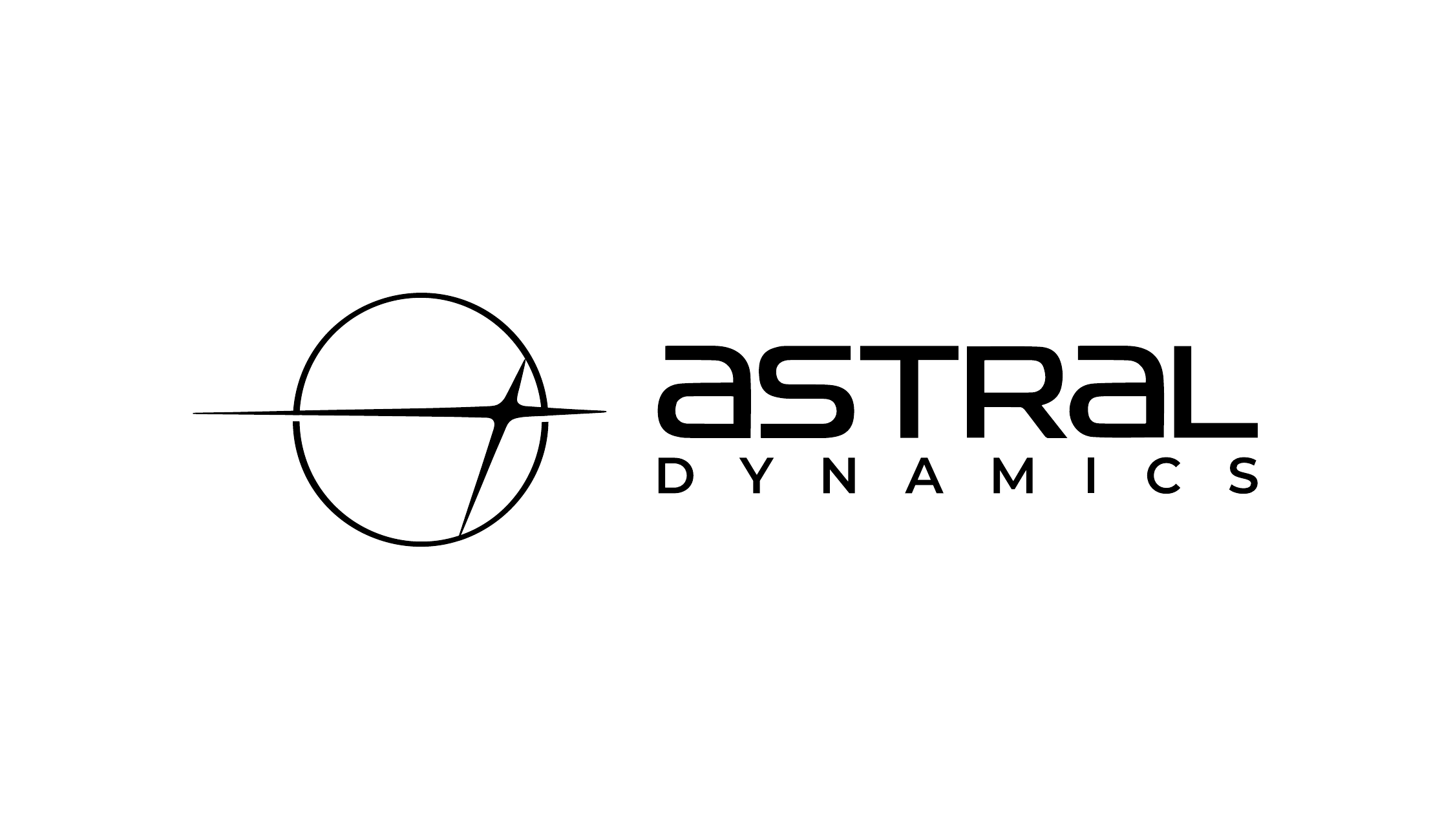}
\end{minipage}\hfill
\begin{minipage}[c]{.5\textwidth}\raggedleft\sffamily\color{chaseblue}\footnotesize\MakeUppercase{Technical Report}\enspace | \enspace November 2026\end{minipage}
\vspace{10mm}
\begin{center}
{\color{chasenavy}\sffamily\bfseries\LARGE CHASE: Curiosity-Guided Skill Acquisition\\[2mm] for Embodied Task Execution\par}
\vspace{5mm}
{\sffamily\small\color{chasenavy}
Zixuan Wang\textsuperscript{1}\quad
Wei Zhang\textsuperscript{2}\quad
Chenlei Zhu\textsuperscript{3}\\[1.5mm]
Naiyu Chen\textsuperscript{3}\quad
Xihua Wang\textsuperscript{1}\quad
Jie Chen\textsuperscript{1,3,*}\par}
\vspace{3mm}
{\sffamily\footnotesize\color{chasenavy}
\textsuperscript{1}Department of Electrical and Computer Engineering, University of Alberta, Edmonton, Canada\\
\textsuperscript{2}Harbin Institute of Technology\\
\textsuperscript{3}School of Biomedical Engineering, Fudan University, China\\[1.5mm]
\textsuperscript{*}Corresponding author\par}
\end{center}
\vspace{5mm}
\noindent\colorbox{chasepaleblue}{\begin{minipage}{\dimexpr\textwidth-2\fboxsep\relax}
\vspace{2mm}\small\textbf{Abstract.}\enspace
Robots executing tasks in partially observed environments must resolve uncertainty while operating with incomplete skills. Inspired by human curiosity, we investigate how robots can formulate task-relevant hypotheses, seek evidence, and acquire missing skills during task execution. We propose CHASE, a Curiosity-guided Harness for Acquiring Skills from Evidence, which operationalizes curiosity as selecting investigations that resolve task uncertainty and expand executable capabilities. Four modules connect dynamic scene understanding, hypothesis-driven planning, on-demand skill acquisition, and verified execution. The planner balances information value, skill reuse, and acquisition/execution cost. Missing skills trigger bounded exploration, learning from verified segments, and held-out admission during acquisition; separate checks verify physical effects and acquired evidence. Newly admitted skills expand the executable investigation set, closing the loop between resolving uncertainty and acquiring capabilities. Simulation and real-robot experiments evaluate investigation, skill acquisition, and task execution. CHASE achieves 50--80\% task success under combined information and capability gaps, with 70--90\% evidence resolution.
\vspace{2mm}\end{minipage}}
\vspace{2mm}
\begin{figure}[!ht]
\centering
\includegraphics[width=0.59\textwidth]{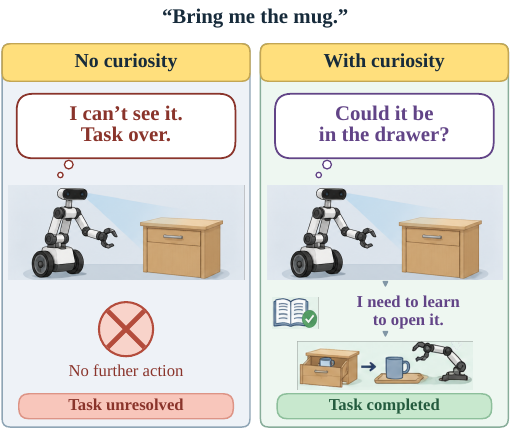}
\caption{A retrieval example of task-directed curiosity. A visibility-only robot acts on current observations. A curious robot proposes locations, seeks evidence, identifies missing capabilities, and acquires validated skills to enable retrieval. The right branch illustrates acquisition mode.}
\label{fig:curiosity-comparison}
\end{figure}
\clearpage
\section{Introduction}
\label{sec:intro}
Humans are naturally curious. Faced with an unfamiliar situation, they formulate hypotheses, seek evidence, and learn new ways to act. This ability motivates a question for embodied intelligence: how can we endow robots with curiosity that supports task execution? A robot may know the desired outcome while lacking information about the environment or a skill needed to achieve it. We study the coupling between these information and capability gaps: useful interactions reveal what the robot needs to know, while acquiring a missing skill can make previously unavailable interactions possible.

A conventional execution pipeline restricted to current observations and a fixed skill library can stall when information or an operation is missing. Figure~\ref{fig:curiosity-comparison} illustrates this issue through a retrieval example. The robot on the left concludes, ``I can't see it. Task over.'' The robot on the right asks, ``Could it be in the drawer?'' It seeks evidence, recognizes the missing opening skill, and acquires that skill before retrieval. This example illustrates a broader principle: \emph{task-directed curiosity} connects task hypotheses, evidence-seeking interactions, and capability expansion, so that uncertainty and missing skills become explicit planning objectives.

Three research directions provide foundations. Semantic maps and scene graphs ground environmental reasoning~\cite{conceptfusion2023,conceptgraphs2023,dovsg2025}, while interactive search reveals hidden targets~\cite{sms2023,himos2023}. Generalist policies support adaptation~\cite{octo2024,openvla2024}, and model-guided data generation supports skill learning~\cite{scalingup2023,robogen2023}. Agentic systems ground and compose operations with planning and feedback~\cite{saycan2022,sayplan2023,innermonologue2022,harnessvla2026}. CHASE connects these directions through an explicit loop: unresolved questions guide investigation and skill acquisition, while newly admitted skills enable further investigations. This coupling requires distinguishing a successful physical operation from an observation that actually answers the question.

We propose \method{}, a \emph{Curiosity-guided Harness for Acquiring Skills from Evidence}, with four modules: dynamic scene understanding, hypothesis-driven curiosity planning, on-demand skill acquisition, and composition with verified execution. Scene memory grounds task hypotheses and retains observation support. A frozen high-level planner ranks investigations by information value, skill reuse, and total cost. Applicable skills execute directly; missing operations trigger bounded generalist exploration during acquisition. Verified segments support local-policy learning, and held-out validation determines skill admission. New skills expand the executable investigation set. Separate physical and evidence checks update the scene and hypotheses, while task-specific predicates verify completion. Held-out reuse evaluation freezes the registry.

The contributions are threefold: \textbf{(1)} a curiosity-driven task-execution framework that jointly ranks investigations by information value, capability-expansion value, and acquisition/execution cost, coupling task questions to skill acquisition and newly enabled investigations; \textbf{(2)} an on-demand skill-acquisition and verification mechanism that connects generalist exploration, feedback-guided refinement, verified-trajectory learning, and held-out admission to form reusable skills, with separate physical and information checks for composition; and \textbf{(3)} a simulation and real-robot experimental evaluation using interactive manipulation tasks to assess task execution, skill reuse, and the coupling between investigation and capability acquisition.

\section{Related Work}
\label{sec:related}
\subsection{Semantic Scene Understanding and Curiosity-driven \mbox{Object Search}}
Semantic representations connect observations to task-relevant entities and relations. Hydra constructs and optimizes hierarchical 3D scene graphs online~\cite{hydra2022}. ConceptFusion~\cite{conceptfusion2023} and VLMaps~\cite{vlmaps2022} fuse visual-language features with spatial maps, enabling open-set queries and language-grounded navigation. ConceptGraphs represents objects and their relations in an open-vocabulary graph~\cite{conceptgraphs2023}; DovSG extends scene-graph reasoning to dynamic updates for long-term mobile manipulation~\cite{dovsg2025}. These representations provide a basis for grounding task hypotheses and preserving observations across interactions.

Finding a hidden target additionally requires deciding what to observe or change. Active perception treats sensing as a controlled process~\cite{bajcsy1988}, while prediction-based intrinsic rewards encourage exploration~\cite{pathak2017}. Goal-oriented semantic exploration~\cite{semexpgoal2020} and VLFM~\cite{vlfm2024} use semantic information to guide search. Semantic Mechanical Search~\cite{sms2023} and HIMOS~\cite{himos2023} incorporate physical interactions with occluders or containers; Explore until Confident uses confidence-calibrated exploration for question answering~\cite{exploreeqa2024}. CHASE builds on these ideas by coupling evidence-seeking requests to capability availability: an informative investigation can trigger acquisition of its missing operation. Physical progress and the evidence needed to resolve a hypothesis are checked separately.

\subsection{Autonomous Robot Skill Acquisition and Adaptation}
Large-scale robot learning supplies increasingly versatile starting capabilities. RT-1~\cite{rt12022} learns language-conditioned control from diverse demonstrations, and RT-2 transfers vision-language knowledge into actions~\cite{rt22023}. Open X-Embodiment studies cross-robot data and policy transfer~\cite{rtx2023}; Octo~\cite{octo2024} and OpenVLA~\cite{openvla2024} provide adaptable generalist policies. Such models offer execution or adaptation backends, while deciding which missing operation to acquire remains a task-level decision.

Foundation models can also guide skill discovery and data generation. MOKA uses mark-based visual prompting for manipulation~\cite{moka2024}; Scaling Up and Distilling Down combines language-guided data collection with policy learning~\cite{scalingup2023}. RoboGen generates tasks, environments, and training supervision~\cite{robogen2023}, while Eureka iteratively improves reward programs~\cite{eureka2023}. RoboCat improves manipulation using demonstrations and self-generated experience~\cite{robocat2023}, and Voyager accumulates executable skills in Minecraft~\cite{voyager2023}. CHASE targets task-triggered capability gaps: bounded generalist attempts produce operation traces, independently verified segments support policy training, and held-out execution determines admission. Diffusion Policy~\cite{diffusionpolicy2023} provides an established learning backend; our focus is the acquisition interface and its feedback to investigation planning.

\subsection{Agent-based Skill Composition and Long-horizon Mobile Manipulation}
Language-model agents connect instructions to executable operations. SayCan grounds skill selection in affordances~\cite{saycan2022}; Code as Policies~\cite{cap2022} and ProgPrompt~\cite{progprompt2022} generate programs over available interfaces. VoxPoser composes spatial value maps for manipulation~\cite{voxposer2023}, and PaLM-E integrates embodied observations with language reasoning~\cite{palme2023}. SayPlan uses hierarchical scene graphs and replanning to scale task planning~\cite{sayplan2023}; Inner Monologue incorporates execution feedback~\cite{innermonologue2022}. Harness VLA orchestrates a frozen contact executor with memory and a fixed primitive library~\cite{harnessvla2026}.

Mobile manipulation further couples semantic decisions with navigation and physical execution. TidyBot combines preference reasoning with household cleanup~\cite{tidybot2023}; Mobile ALOHA learns bimanual whole-body behaviors from teleoperation~\cite{mobilealoha2024}. AutoRT uses foundation models to orchestrate robot data collection~\cite{autort2024}. These systems motivate composition, feedback, and reusable interfaces across extended tasks. CHASE investigates a complementary coupling: the agent schedules investigations and existing skills, requests acquisition when capability is insufficient, and revisits pending requests after admission. Our comparisons therefore share initial capabilities and budgets, separating orchestration choices from changes in the skill registry.

\begin{figure}[!tbp]
\centering
\includegraphics[width=\linewidth]{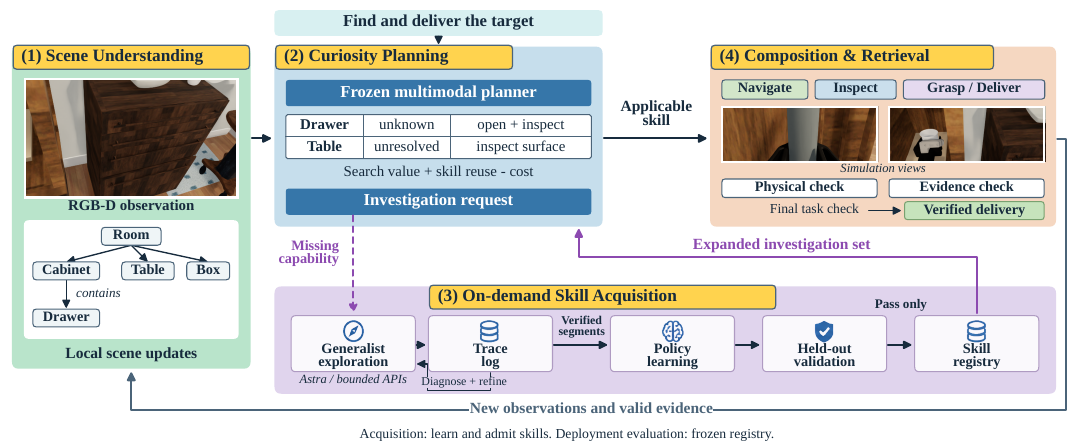}
\caption{Retrieval instantiation of the \method{} pipeline. Scene memory grounds task hypotheses, shown here as candidate locations; requests reuse applicable skills or trigger exploration, verified-trace learning, and held-out admission. Admitted skills expand subsequent investigations. Independent physical and evidence checks feed local updates until released delivery is verified. Insets show simulation diagnostics. Acquisition enables learning; held-out reuse freezes the registry.}
\label{fig:framework}
\end{figure}

\section{Method}
\label{sec:method}
\subsection{Problem Formulation and System State}
\label{sec:problem}
Given instruction $\mathcal{L}$, observation history $O_{0:t}$, and an initially incomplete skill library $\mathcal{S}_0$, the robot seeks task completion under partial observability. Observations contain RGB-D and robot state. Let $G_{\mathcal{L}}$ denote task-specific completion, whose predicates can depend on object location, state, content, or interaction outcomes. We seek
\begin{equation}
\max_{\Pi}\;P_{\Pi}\!\left(G_{\mathcal{L}}=1\mid
\mathcal{L},O_{0:t},\mathcal{S}_t\right),\quad C_{0:T}\le B,
\label{eq:objective}
\end{equation}
where $\Pi$ selects investigations, existing skills, and acquisition attempts; $C$ includes sensing, execution, failed attempts, and skill acquisition. Hidden simulator state supplies initialization and offline scoring only.

The system maintains a scene graph $\mathcal{G}_t$, task-conditioned hypotheses $\mathcal{H}_t$, evidence memory $\mathcal{E}_t$, and a skill registry $\mathcal{S}_t$. Hypotheses concern uncertain task facts or operation preconditions. Scene memory describes the environment; questions identify what must be established next. Figure~\ref{fig:framework} instantiates the framework through retrieval, whose goal requires the correct object, any requested content, and released delivery. Investigations expose missing skills; admitted skills expand executable investigations.

\subsection{Stage 1: Dynamic 3D Scene Understanding}
Following DovSG~\cite{dovsg2025}, posed RGB-D observations support open-vocabulary instance association, three-dimensional geometry fusion, spatial relations, and local updates. We represent $\mathcal{G}_t=(\mathcal{V}_t,\mathcal{R}_t)$, where nodes describe objects, containers, and observed regions, and edges encode relations such as \texttt{on}, \texttt{inside}, and \texttt{near}. New views are aligned to the existing map before association and fusion. Interactions trigger updates to affected instances and relations, preserving unrelated scene memory.

We augment this representation with articulation estimates, visible support, observation history, candidate interaction attributes, and reachability evidence. These are task-specific extensions; an object label alone does not establish that it is openable or reachable. Geometry and kinematics provide candidate checks, while execution supplies physical verification. Each attribute retains its observation source, camera pose, timestamp, and validity scope. Moving an occluder invalidates dependent visibility claims; opening a different drawer need not invalidate an already observed surface.

\subsection{Stage 2: Hypothesis-Driven Curiosity Planning}
A frozen multimodal planner combines $\mathcal{L}$, $\mathcal{G}_t$, and evidence to propose hypotheses about location, contents, object state, or feasible interaction conditions. Each $h\in\mathcal{H}_t$ binds a proposition to scene entities, evidence references, and an investigation. Its status is \emph{unobserved}, \emph{inconclusive}, \emph{supported}, or \emph{contradicted}. Retrieval includes candidate locations and an \emph{elsewhere} alternative. Missing detections exclude a region only with adequate visibility, coverage, and detector reliability; unsupported or stale observations leave the question unresolved.

An investigation request is
\begin{equation}
u=(h,q,g,s_{\mathrm{req}},\widehat y,\kappa),
\label{eq:request}
\end{equation}
where $q$ is the unresolved question, $g$ the operation goal, $s_{\mathrm{req}}$ the required skill, $\widehat y$ the expected evidence, and $\kappa$ the capability status. For a closed drawer, the request may bind a location hypothesis to ``inspect interior,'' require opening, and identify that no applicable opening skill is available.

For unresolved questions, we jointly rank executable investigations and budget-feasible acquisition requests:
\begin{equation}
u_t^*=\arg\max_{u\in\mathcal{U}_t(B)}
\left[\widehat V_{\mathrm{info}}(u)
+\eta\widehat V_{\mathrm{skill}}(u)-\lambda\widehat C(u)\right].
\label{eq:score}
\end{equation}
For question $q$, let $b_t^q$ be a normalized belief over a finite answer set $\mathcal{A}_q$. Information value is expected entropy reduction,
\begin{equation}
\widehat V_{\mathrm{info}}(u)=
\frac{H(b_t^q)-\sum_y\widehat p(y\mid u)H(\widehat b_t^{q,u,y})}
{\max(\log|\mathcal{A}_q|,\epsilon)}.
\label{eq:ig}
\end{equation}
For location questions, answers are candidate regions plus elsewhere; a content question can use empty/nonempty. Outcomes $y$ support, contradict, or leave an answer inconclusive. Failed or insufficient observations preserve the belief unless they provide valid new evidence. The outcome model groups acquisition observations by question type, operation, and visibility, with conservative back-off for unsupported cases. Semantic rankings require calibration before use as probabilities; otherwise a uniform answer prior is retained.

Skill value is the predicted weighted fraction of pending, currently unserved requests that the required operation could enable. It is a potential-coverage estimate, not proof of skill generalization. Actual expansion is checked after admission. Cost includes execution, exploration, learning, validation, and failure recovery, normalized by the remaining budget. Thus a missing skill does not remove a useful investigation; its acquisition cost competes with accessible alternatives. A finite budget and bounded retries prevent indefinite learning on one request. The scoring estimators and weights are fixed after acquisition/validation and remain unchanged during controlled reuse evaluation.

Let $\mathcal{I}(\mathcal{G}_t,\mathcal{S}_t)$ denote currently executable investigations under validated applicability. The capability feedback from admitting $s$ is
\begin{equation}
\Delta\mathcal{I}_s=
\mathcal{I}(\mathcal{G}_t,\mathcal{S}_t\cup\{s\})
\setminus\mathcal{I}(\mathcal{G}_t,\mathcal{S}_t).
\label{eq:expansion}
\end{equation}
Planning is repeated with the updated set. An empty expansion is possible; registration alone does not prove increased capability.

\subsection{Stage 3: On-Demand Autonomous Skill Acquisition}
\label{sec:learning}
\paragraph{Skill retrieval and applicability}
Each registry entry specifies typed inputs/outputs, preconditions, executor, expected physical effects, termination/timeout, success checker, validated scope, and provenance. Retrieval requires both a matching operation and supported current conditions. If a stored skill is inapplicable, the planner first considers preparatory navigation, viewpoint, or environment changes. An unsatisfied capability triggers acquisition only when its budget and permitted interfaces support an attempt.

\paragraph{Exploration and refinement}
A frozen generalist, such as Astra, receives the operation goal, RGB-D, robot state, allowed APIs, and prior attempt feedback. Its bounded rollout is
\begin{equation}
\tau^{(k)}\sim\pi_{\mathrm{gen}}
(\,\cdot\mid g,O_{0:t},\mathcal{A},d^{(<k)}),
\label{eq:exploration}
\end{equation}
where $d^{(<k)}$ records earlier failure stages and diagnoses. Each attempt logs observations, commands, robot and object-state estimates, termination, and available contact/force feedback. Diagnosis conditions the next attempt without updating generalist weights. This in-context refinement is distinct from learning a standalone policy.

\paragraph{Policy formation and admission}
Operation boundaries define training segments with pre-contact context and terminal verification. A segment enters positive imitation data only if its own physical postcondition is verified, even when the enclosing task later fails. Failed or uncertain segments remain diagnostic/applicability data; they are not relabeled as successful actions. A goal-conditioned local policy is fitted from verified segments using an established imitation-learning backend, e.g., diffusion policy~\cite{diffusionpolicy2023}. This work specifies data admission and invocation contracts rather than a new policy architecture.

The candidate is evaluated on disjoint validation configurations spanning object geometry, start pose, articulation, and viewpoint. Predeclared success, duration, and execution-failure criteria determine admission; failed validation returns the candidate to refinement or rejects it. Only admitted policies enter $\mathcal{S}_{t+1}$ with dataset lineage and supported conditions. Exploration attempts also return the checks in~\eqref{eq:dual}, immediately updating scene and evidence memory. A resolved request need not wait for policy admission; admitted skills enable subsequent requests. Generalist success and reusable-skill admission are distinct outcomes.

\subsection{Stage 4: Skill Composition and Verified Task Execution}
For the selected request, the harness composes applicable navigation, viewing, interaction, and manipulation skills. Each call returns separate physical and information outcomes,
\begin{equation}
(v^{\mathrm{phys}},v^{\mathrm{info}})
=\left(V_s^{\mathrm{phys}}(\tau),
V_q^{\mathrm{info}}(\tau,\mathcal{E}_t)\right).
\label{eq:dual}
\end{equation}
Physical checks verify effects such as opening, retention, or placement; information checks verify observation support for the task question, including state, content, or coverage-qualified location conclusions. A physically opened but still occluded drawer leaves its content unresolved. Conversely, a failed grasp can preserve valid location evidence.

Receipts update affected geometry, invalidate dependent evidence, and update hypotheses from valid observations. Unresolved questions return to Stage 2. With required preconditions supported, the harness executes goal-directed operations without requiring further information gain. Failure invokes bounded recovery, re-investigation, or acquisition. Completion checks $G_{\mathcal{L}}$ independently; retrieval requires released delivery, beyond detection or possession.

\paragraph{Learning and evaluation boundary}
Acquisition episodes enable exploration, learning, and registry updates, accounting for their full cost. Admission uses disjoint validation configurations. A separate held-out reuse evaluation freezes the planner, policies, registry, estimators, and thresholds; only episode evidence changes online. Acquisition evaluates capability formation; frozen reuse evaluates transfer and composition. Held-out reuse trajectories do not train or admit skills. Recorded diagnostics with authored executors remain separate from learned-policy evaluation.

\begin{figure}[!tbp]
\centering
\includegraphics[width=\linewidth]{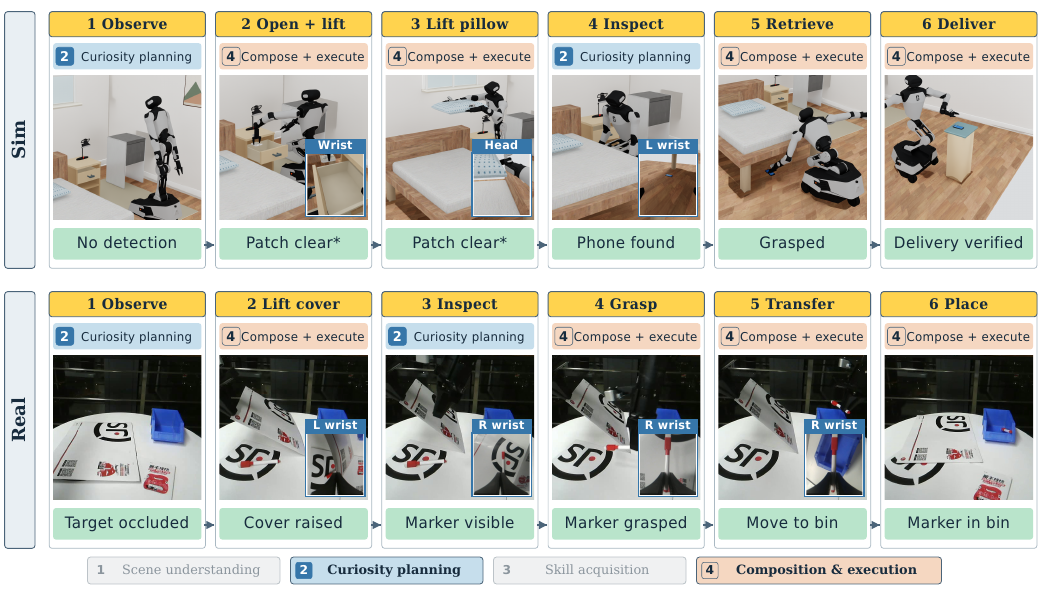}
\caption{Recorded simulation and real-robot sequences. Sim: phone investigation and retrieval with existing skills; *clear only within the observed patch. Real: lift an occluding board, reveal and grasp a marker, and place it in a bin; wrist insets show the same video frame. Module bars indicate functional roles: planning (2) and composition/execution (4); other modules are muted.}
\label{fig:experimental-setup}
\end{figure}

\section{Experiments}
\label{sec:diagnostics}
We independently vary task uncertainty and missing operations to evaluate \textbf{Q1}, evidence-seeking investigation; \textbf{Q2}, reusable skill acquisition; \textbf{Q3}, capability feedback; and \textbf{Q4}, verified task composition. Ranges denote the minimum and maximum measured performance across scenes or configurations, except the weighted summary All; Table~\ref{tab:skill-results} reports observed skill-acquisition bounds. Bold marks favorable range endpoints; arrows indicate preferred direction. Calls count high-level invocations per episode, including failed attempts.

\subsection{Experimental Setup and Task Design}
\label{sec:evaluation}
\subsubsection{Platforms and observations}
Simulation uses BEHAVIOR-1K~\cite{behavior1k2024} with a Galaxea R1 Pro robot. Table~\ref{tab:task-families} lists the six evaluated scene instances: a bedroom, drawers, a game room, kitchen and living-room cabinets, and a lidded container. Physical experiments use an AgileX Split Aloha for hard-drive search, office-object search, and container-interior inspection in clutter. Figure~\ref{fig:experimental-setup} shows simulation phone retrieval and real-robot occluder lifting and marker retrieval. Matched comparisons share RGB-D observations, robot state, and permitted control APIs. Hidden-target annotations used for scoring are excluded from the agent's observations.

\subsubsection{Scene instances and experimental conditions}
Scene identity is separate from the information and capability conditions. \textbf{T1} withholds the target location while retaining all required skills; \textbf{T2} supplies the location but withholds a required operation; \textbf{T3} combines unknown location and a missing operation. A paired control supplies the location and retains the required skills. \textbf{T4} denotes long-horizon composition and is an additional episode attribute, which can overlap T1--T3. It is not a fourth mutually exclusive information/capability condition.

For mug retrieval, T1 hides the cabinet location, T2 withholds opening, and T3 combines both gaps. A supplied location provides no oracle grasp or motion. Each episode records the scene, condition, and initial library; acquisition additionally requires attempts and admission records. The evaluation comprises 120 T1 trials and 60 each for T2 and T3 per method: 1,440 simulation trials across six methods.

\begin{table}[t]
\centering
\caption{Simulation scene instances: 20 trials per scene--condition pair and method. All scenes use T1; S2, S4, and S6 additionally use T2 and T3 (240 trials per method).}
\label{tab:task-families}
\footnotesize
\setlength{\tabcolsep}{3pt}
\begin{tabular}{>{\raggedright\arraybackslash}p{.18\columnwidth}>{\raggedright\arraybackslash}p{.75\columnwidth}}
\toprule
Scene & Investigation and retrieval case\\
\midrule
S1: Bedroom & Multi-Region Sequential Investigation and Retrieval (phone)\\[1mm]
S2: Drawers & Multi-Compartment Investigation and Key Retrieval\\[1mm]
S3: Game room & Multi-Compartment Investigation and Toy-Car Retrieval\\[1mm]
S4: Kitchen & Cabinet-Interior Investigation and Mug Retrieval\\[1mm]
S5: Living room & Cabinet-Interior Investigation and Figurine Retrieval\\[1mm]
S6: Container & Occluder Removal and Container-Interior Investigation (toy car)\\
\bottomrule
\end{tabular}
\end{table}

\subsubsection{Acquisition, admission, and reuse}
Acquisition starts from $\mathcal{S}_0$ and permits bounded exploration, learning, and registry updates. Admission and frozen-policy reuse use disjoint configurations with changed layouts, unseen objects, or combined shifts; scene and evidence updates remain enabled. Baselines use matched initial observations, initial capabilities, permitted APIs, and per-stage resource limits; all attempts, validation executions, and failed candidates enter acquisition cost.

\begin{table}[t]
\centering
\caption{Simulation performance. Ranges are scene/configuration minima--maxima; All weights condition endpoints by 0.5:0.25:0.25. Time includes acquisition; Search + Acquisition omits capability-value ranking.}
\label{tab:main-results}
\footnotesize
\setlength{\tabcolsep}{2pt}
\begin{tabular}{>{\raggedright\arraybackslash}p{.30\columnwidth}cccc}
\toprule
Method & \shortstack{T1\\(\%) $\uparrow$} & \shortstack{T2\\(\%) $\uparrow$} & \shortstack{T3\\(\%) $\uparrow$} & \shortstack{All\\(\%) $\uparrow$}\\
\midrule
SSF (adapted) & 60--80 & 0--10 & 0--10 & 30--45\\[1mm]
Generalist Only & 50--70 & 25--50 & 20--40 & 36.25--57.5\\[1mm]
Reactive, Fixed Skills & 20--40 & 0--10 & 0--10 & 10--25\\[1mm]
Search + Fixed Skills & 65--85 & 0--10 & 0--10 & 32.5--47.5\\[1mm]
Search + Acquisition & 65--85 & 55--80 & 45--70 & 57.5--80\\[1mm]
\textbf{Full CHASE (ours)} & \textbf{70--90} & \textbf{60--85} & \textbf{50--80} & \textbf{62.5--86.25}\\[1mm]
\bottomrule
\end{tabular}\par
\vspace{2mm}
\begin{tabular}{>{\raggedright\arraybackslash}p{.37\columnwidth}ccc}
\toprule
Method & \shortstack{Locate\\(\%) $\uparrow$} & \shortstack{Calls\\(count) $\downarrow$} & \shortstack{Time\\(s) $\downarrow$}\\
\midrule
SSF (adapted) & 60--80 & 10--18 & 180--360\\[1mm]
Generalist Only & 55--75 & 14--24 & 300--600\\[1mm]
Reactive, Fixed Skills & 25--45 & \textbf{5--10} & \textbf{90--210}\\[1mm]
Search + Fixed Skills & 65--85 & 9--16 & 180--330\\[1mm]
Search + Acquisition & 70--90 & 12--22 & 600--2100\\[1mm]
\textbf{Full CHASE (ours)} & \textbf{75--95} & 10--20 & 600--1800\\[1mm]
\bottomrule
\end{tabular}
\end{table}

\subsection{Baselines and End-to-End Task Performance}
\subsubsection{Comparison methods}
\textbf{SSF (adapted)}~\cite{ssf2026} is the external search baseline. It fuses semantic and geometric evidence in a voxel belief map and selects actions by expected revealed belief mass using a fixed skill library. The adaptation retains these updates and ranking while using the common robot executors. T1 matches available skills to compare search; in T2--T3, fixed-library failures identify capability limits, not search quality alone. Fixed-skill success in T2--T3 is limited to 0--10\%; without a library alternative, a withheld necessary operation prevents completion. These outcomes characterize the evaluated library limitations and do not demonstrate skill acquisition. \textbf{Generalist Only} retains investigation but uses a frozen generalist without policy learning or registration. \textbf{Reactive, Fixed Skills} acts on current observations without hypothesis-driven investigation. \textbf{Search + Fixed Skills} retains hypotheses and evidence memory but disables acquisition. \textbf{Search + Acquisition} enables learning and admission with $\eta=0$. \textbf{Full CHASE} uses the complete framework. These variants isolate investigation, learning, and capability-value ranking; Harness VLA~\cite{harnessvla2026} is not an additional reproduced baseline.

Retrieval requires the correct target, requested content conditions, and stable released delivery; search-only tasks require supported localization. Localization is also scored separately. Table~\ref{tab:main-results} isolates search (T1), acquisition (T2), and their coupling (T3). The task mix is T1:T2:T3 = 120:60:60. All is the weighted envelope of the T1--T3 endpoints, giving 62.5--86.25\% for CHASE; it is not a pooled success rate or an observed aggregate minimum--maximum. Composition is assessed within these conditions without double-counting T4-tagged episodes. High-level calls, execution time, and wall-clock time are evaluated over all episodes. The acquisition-enabled comparison includes the cost of learning rather than reporting only the speed of already learned executors.

The capability gap separates the methods most clearly: CHASE achieves 60--85\% success in T2 and 50--80\% in T3, while fixed-library methods remain at 0--10\%. Search + Acquisition achieves 55--80\% and 45--70\%, respectively. These comparisons connect acquisition and capability-value ranking to completion when required operations are missing. T1 evaluates investigation with all required skills available; differences among acquisition variants in T1 are not attributed to learning new skills.

First-task time includes exploration, data preparation, training, admission, and execution. CHASE requires 600--1800\,s across the evaluated task mix; frozen reuse is assessed separately after acquisition, with no reuse trajectories used for training or admission.

\subsection{Hypothesis-Driven Investigation}
Q1 links a hypothesis, investigation, observation, and evidence update. With fixed skills, hypothesis-driven search achieves 65--85\% T1 success, compared with 20--40\% for reactive execution. This comparison supports explicit investigation when capabilities are sufficient. In T3, removing information ranking changes evidence resolution from 70--90\% to 60--85\% (Table~\ref{tab:ablation-results}), linking investigation selection to supported answers.

Figure~\ref{fig:experimental-setup} illustrates this process: nondetection motivates further investigation, exposed patches provide local evidence, and an under-bed observation supports phone localization before retrieval. A clear patch does not exclude a whole occluded region. False exclusion means ruling out a region without adequate evidence; false completion means declaring success with an unsatisfied terminal predicate. Absent-target episodes are scored separately from positive retrieval.

\subsection{On-Demand Skill Acquisition and Reuse}
\subsubsection{From exploration to admitted skills}
Q2 evaluates independently callable policy formation. Operation records retain applicability failures, generalist attempts, diagnoses, verified segments, and admission outcomes. Table~\ref{tab:skill-results} reports the acquired operations and performance bounds for exploration, independent validation, and frozen reuse. Generalist success may resolve the current question before admission. Rejected candidates and failed attempts remain recorded.

\begin{table}[t]
\centering
\caption{Skill-acquisition results: lower bounds on exploration, independent validation, and frozen reuse success; upper bounds on total acquisition time.}
\label{tab:skill-results}
\footnotesize
\setlength{\tabcolsep}{2.5pt}
\begin{tabular}{>{\raggedright\arraybackslash}p{.40\columnwidth}cccc}
\toprule
Operation & \shortstack{Explore\\(\%) $\uparrow$} & \shortstack{Valid.\\(\%) $\uparrow$} & \shortstack{Reuse\\(\%) $\uparrow$} & \shortstack{Time\\(min) $\downarrow$}\\
\midrule
Drawer/cabinet opening & $\geq$60 & $\geq$80 & $\geq$75 & $\leq$30\\[1mm]
Occluder lift and restore & $\geq$55 & $\geq$80 & $\geq$70 & $\leq$40\\[1mm]
Lid removal and return & $\geq$60 & $\geq$80 & $\geq$75 & $\leq$30\\[1mm]
\bottomrule
\end{tabular}
\end{table}

\subsubsection{Generalization and acquisition efficiency}
Held-out reuse varies geometry, articulation, starting configuration, or viewpoint within the validated skill scope. Table~\ref{tab:skill-results} addresses capability formation directly: validation success is at least 80\% for all three operations, while frozen reuse success is at least 75\% for opening and lid handling and 70\% for occluder handling. These outcomes establish independent policy reuse beyond a successful generalist attempt. Acquisition takes at most 30 minutes for opening and lid handling and 40 minutes for occluder handling. The cost accompanies the acquired capability and must be retained when assessing first-task efficiency.

\subsection{Capability Feedback and Ablation Studies}
Q3 connects acquired operations to the executable investigation set in~\eqref{eq:expansion}. Within validated applicability, opening enables closed-interior inspection, occluder lifting enables covered-region inspection, and lid removal enables container inspection. Table~\ref{tab:skill-results} establishes reuse of these operations; T3 evaluates their combination with investigation when both location and operation are initially unavailable.

The $\eta=0$ comparison isolates capability-value ranking: CHASE reports 50--80\% T3 success versus 45--70\% for Search + Acquisition. Information ranking, refinement, applicability gating, and separate outcome checks are examined in Table~\ref{tab:ablation-results}. Merging physical and information checks gives 50--80\% evidence resolution versus 70--90\% for the full system, supporting their separate treatment. Without refinement feedback, success is 35--70\% and time is 900--2400\,s. Together, skill-level reuse and task-level comparisons evaluate how admitted operations support later investigations within validated skill scopes. Decision comparisons match scene seeds and budgets; data-collection effects are distinguished from invocation effects.

\begin{table}[t]
\centering
\caption{T3 ablations. Ranges are minima--maxima across scenes/configurations. Evidence is the resolved-question rate; time includes acquisition. Libraries and budgets are matched.}
\label{tab:ablation-results}
\footnotesize
\setlength{\tabcolsep}{2.5pt}
\begin{tabular}{>{\raggedright\arraybackslash}p{.47\columnwidth}ccc}
\toprule
Variant & \shortstack{Success\\(\%) $\uparrow$} & \shortstack{Evidence\\(\%) $\uparrow$} & \shortstack{Time\\(s) $\downarrow$}\\
\midrule
Full CHASE & \textbf{50--80} & \textbf{70--90} & \textbf{600--1800}\\[1mm]
Without information ranking & 40--75 & 60--85 & 720--2100\\[1mm]
Without refinement feedback & 35--70 & 55--80 & 900--2400\\[1mm]
Without applicability gating & 35--70 & 55--85 & 720--2400\\[1mm]
Merged outcome checks & 35--70 & 50--80 & 600--2100\\[1mm]
\bottomrule
\end{tabular}
\end{table}

\subsection{Real-Robot Execution in Clutter}
Q4 evaluates three Split Aloha tasks in naturally cluttered workspaces. \textbf{R1: Occluded Hard-Drive Search} uses the recorded configurations beneath a magazine and behind objects. \textbf{R2: Office Object Search} adds a target such as a pen among office items, testing sequential investigation of accessible candidate regions. \textbf{R3: Container-Interior Inspection} adds a target-presence question about a container, requiring an adequate interior observation and a supported answer. All three tasks use the same comparison methods and instruction-specific completion checks. The Real row of Fig.~\ref{fig:experimental-setup} illustrates an R2 office-object case: lifting the board exposes a marker, a wrist view supports localization, and grasping and placement complete retrieval. The module bars connect investigation (2) to physical execution (4); quantitative outcomes appear in Table~\ref{tab:real-results}.

\begin{table}[t]
\centering
\caption{Real-robot performance. Ranges are minima--maxima across configurations. HDD: hard-drive search; Pen: office search; Inspect: container inspection. Time excludes training and admission.}
\label{tab:real-results}
\footnotesize
\setlength{\tabcolsep}{2.5pt}
\begin{tabular}{>{\raggedright\arraybackslash}p{.32\columnwidth}ccccc}
\toprule
Method & \shortstack{HDD\\(\%) $\uparrow$} & \shortstack{Pen\\(\%) $\uparrow$} & \shortstack{Inspect\\(\%) $\uparrow$} & \shortstack{Calls\\(count) $\downarrow$} & \shortstack{Time\\(s) $\downarrow$}\\
\midrule
SSF (adapted) & 45--70 & 40--65 & 35--60 & 10--18 & 150--360\\[1mm]
Generalist Only & 40--65 & 35--60 & 35--65 & 14--24 & 240--600\\[1mm]
Reactive, Fixed Skills & 15--40 & 15--40 & 10--35 & \textbf{5--10} & \textbf{60--180}\\[1mm]
Search + Fixed Skills & 50--75 & 45--70 & 40--65 & 9--17 & 150--360\\[1mm]
Search + Acquisition & 55--80 & 50--75 & 50--80 & 12--22 & 240--600\\[1mm]
\textbf{Full CHASE (ours)} & \textbf{60--85} & \textbf{55--80} & \textbf{55--85} & 10--20 & 210--540\\[1mm]
\bottomrule
\end{tabular}
\end{table}

R1 and R2 require supported target localization; R3 requires a correct target-presence answer backed by adequate interior visibility; the same verification wrapper is used for SSF. Failure to see inside is inconclusive, not evidence of absence. Grasping, delivery, or opening is required only when specified by the actual task configuration. The initial registry determines whether an interaction uses an existing skill or triggers acquisition; a successful container inspection alone does not establish learning.

Table~\ref{tab:real-results} tests the investigation and verification loop on hardware. CHASE achieves 60--85\% hard-drive search success, 55--80\% office-object search success, and 55--85\% container-inspection success; SSF achieves 45--70\%, 40--65\%, and 35--60\%, respectively. These results address supported localization and inspection in physical clutter. Both rows of Fig.~\ref{fig:experimental-setup} illustrate investigation followed by grasping and placement. The real case shows how changing visibility enables retrieval, while physical success predicates remain instruction-specific.

Calls and execution time include failures but exclude training and admission. First-task wall-clock time must additionally include any triggered acquisition cost from Table~\ref{tab:skill-results}. Resets and interventions are distinguished from autonomous completion.

\subsection{Failure Modes and Scope}
The ablations identify sensitivity to several failure mechanisms. Removing applicability gating gives 35--70\% T3 success versus 50--80\% for CHASE, consistent with the importance of matching a skill to supported conditions. Merged checks reduce the reported evidence-resolution range, showing why physical progress alone is insufficient to establish an answer.

Failure attribution distinguishes unsupported hypotheses, invalid evidence updates, missing or misapplied skills, execution failures, and terminal-verification errors. A failed opening leaves an operation gap; an opened but unobserved interior leaves an information gap. Correct localization followed by a failed grasp remains a manipulation failure, and possession without released delivery does not satisfy retrieval. These distinctions connect investigation, acquisition, and verification to separate stages of task completion. Conclusions concern the evaluated scenes, skill scopes, and budgets; physical search and inspection do not by themselves establish broader manipulation coverage.

\section{Conclusion}
\label{sec:discussion}
\method{} couples evidence-seeking investigation, reusable skill acquisition, and verified task composition. Fixed-library search supports T1 investigation, while CHASE achieves 60--85\% success in T2 and 50--80\% in T3 when capabilities are incomplete. Frozen reuse reaches at least 70\% for all acquired operations, and separate outcome checks support 70--90\% evidence resolution in T3. Together, these results connect capability formation to investigation and task execution. Physical search and inspection achieve 55--85\% success across three task types. Remaining challenges include reliable evidence under incomplete visibility, acquisition-cost estimation, and validation across broader objects and environments.
\begingroup
\small

\endgroup

\begin{thebibliography}{10}
\providecommand{\url}[1]{#1}
\csname url@samestyle\endcsname
\providecommand{\newblock}{\relax}
\providecommand{\bibinfo}[2]{#2}
\providecommand{\BIBentrySTDinterwordspacing}{\spaceskip=0pt\relax}
\providecommand{\BIBentryALTinterwordstretchfactor}{4}
\providecommand{\BIBentryALTinterwordspacing}{\spaceskip=\fontdimen2\font plus
\BIBentryALTinterwordstretchfactor\fontdimen3\font minus
  \fontdimen4\font\relax}
\providecommand{\BIBforeignlanguage}[2]{{%
\expandafter\ifx\csname l@#1\endcsname\relax
\typeout{** WARNING: IEEEtran.bst: No hyphenation pattern has been}%
\typeout{** loaded for the language `#1'. Using the pattern for}%
\typeout{** the default language instead.}%
\else
\language=\csname l@#1\endcsname
\fi
#2}}
\providecommand{\BIBdecl}{\relax}
\BIBdecl

\bibitem{conceptfusion2023}
K.~M. Jatavallabhula \emph{et~al.}, ``{ConceptFusion: Open-set multimodal 3D
  mapping},'' in \emph{Proc. Robot.: Sci. Syst. (RSS)}, 2023.

\bibitem{conceptgraphs2023}
Q.~Gu \emph{et~al.}, ``{ConceptGraphs: Open-Vocabulary 3D Scene Graphs for
  Perception and Planning},'' in \emph{Proc. IEEE Int. Conf. Robot. Autom.
  (ICRA)}, 2024, pp. 5021--5028.

\bibitem{dovsg2025}
Z.~Yan \emph{et~al.}, ``{Dynamic Open-Vocabulary 3D Scene Graphs for Long-Term
  Language-Guided Mobile Manipulation},'' \emph{IEEE Robot. Autom. Lett.},
  vol.~10, no.~5, pp. 4252--4259, 2025.

\bibitem{sms2023}
S.~Sharma \emph{et~al.}, ``{Semantic Mechanical Search with Large Vision and
  Language Models},'' in \emph{Proc. Conf. Robot Learn. (CoRL)}, 2023, pp.
  971--1005.

\bibitem{himos2023}
F.~Schmalstieg, D.~Honerkamp, T.~Welschehold, and A.~Valada, ``{Learning
  Hierarchical Interactive Multi-Object Search for Mobile Manipulation},''
  \emph{IEEE Robot. Autom. Lett.}, vol.~8, no.~12, pp. 8549--8556, 2023.

\bibitem{octo2024}
D.~Ghosh \emph{et~al.}, ``{Octo: An Open-Source Generalist Robot Policy},'' in
  \emph{Proc. Robot.: Sci. Syst. (RSS)}, 2024.

\bibitem{openvla2024}
M.~J. Kim \emph{et~al.}, ``{OpenVLA: An Open-Source Vision-Language-Action
  Model},'' in \emph{Proc. Conf. Robot Learn. (CoRL)}, 2025, pp. 2679--2713.

\bibitem{scalingup2023}
H.~Ha, P.~Florence, and S.~Song, ``{Scaling Up and Distilling Down:
  Language-Guided Robot Skill Acquisition},'' in \emph{Proc. Conf. Robot Learn.
  (CoRL)}, 2023, pp. 3766--3777.

\bibitem{robogen2023}
Y.~Wang \emph{et~al.}, ``{{R}obo{G}en: Towards Unleashing Infinite Data for
  Automated Robot Learning via Generative Simulation},'' in \emph{Proc. Int.
  Conf. Mach. Learn. (ICML)}, 2024, pp. 51\,936--51\,983.

\bibitem{saycan2022}
B.~Ichter \emph{et~al.}, ``{Do As I Can, Not As I Say: Grounding Language in
  Robotic Affordances},'' in \emph{Proc. Conf. Robot Learn. (CoRL)}, 2023, pp.
  287--318.

\bibitem{sayplan2023}
K.~Rana, J.~Haviland, S.~Garg, J.~Abou-Chakra, I.~Reid, and N.~Suenderhauf,
  ``{SayPlan: Grounding Large Language Models using 3D Scene Graphs for
  Scalable Robot Task Planning},'' in \emph{Proc. Conf. Robot Learn. (CoRL)},
  2023, pp. 23--72.

\bibitem{innermonologue2022}
W.~Huang \emph{et~al.}, ``{Inner Monologue: Embodied Reasoning through Planning
  with Language Models},'' in \emph{Proc. Conf. Robot Learn. (CoRL)}, 2023, pp.
  1769--1782.

\bibitem{harnessvla2026}
Y.~Zhang \emph{et~al.}, ``{Harness VLA: Steering Frozen VLAs into Reliable
  Manipulation Primitives via Memory-Guided Agents},'' arXiv preprint, 2026,
  arXiv:2607.08448.

\bibitem{hydra2022}
N.~Hughes, Y.~Chang, and L.~Carlone, ``{Hydra: A Real-time Spatial Perception
  System for 3D Scene Graph Construction and Optimization},'' in \emph{Proc.
  Robot.: Sci. Syst. (RSS)}, 2022.

\bibitem{vlmaps2022}
C.~Huang, O.~Mees, A.~Zeng, and W.~Burgard, ``{Visual Language Maps for Robot
  Navigation},'' in \emph{Proc. IEEE Int. Conf. Robot. Autom. (ICRA)}, 2023,
  pp. 10\,608--10\,615.

\bibitem{bajcsy1988}
R.~Bajcsy, ``{Active perception},'' \emph{Proc. IEEE}, vol.~76, no.~8, 1988.

\bibitem{pathak2017}
D.~Pathak, P.~Agrawal, A.~A. Efros, and T.~Darrell, ``{Curiosity-driven
  Exploration by Self-supervised Prediction},'' in \emph{Proc. Int. Conf. Mach.
  Learn. (ICML)}, 2017, pp. 2778--2787.

\bibitem{semexpgoal2020}
D.~S. Chaplot, D.~Gandhi, A.~Gupta, and R.~Salakhutdinov, ``{Object Goal
  Navigation using Goal-Oriented Semantic Exploration},'' arXiv preprint, 2020,
  arXiv:2007.00643.

\bibitem{vlfm2024}
N.~Yokoyama, S.~Ha, D.~Batra, J.~Wang, and B.~Bucher, ``{VLFM: Vision-Language
  Frontier Maps for Zero-Shot Semantic Navigation},'' in \emph{Proc. IEEE Int.
  Conf. Robot. Autom. (ICRA)}, 2024, pp. 42--48.

\bibitem{exploreeqa2024}
A.~Z. Ren, J.~Clark, A.~Dixit, M.~Itkina, A.~Majumdar, and D.~Sadigh,
  ``{Explore until Confident: Efficient Exploration for Embodied Question
  Answering},'' in \emph{Proc. Robot.: Sci. Syst. (RSS)}, 2024.

\bibitem{rt12022}
A.~Brohan \emph{et~al.}, ``{RT-1: Robotics Transformer for Real-World Control
  at Scale},'' arXiv preprint, 2022, arXiv:2212.06817.

\bibitem{rt22023}
B.~Zitkovich \emph{et~al.}, ``{RT-2: Vision-Language-Action Models Transfer Web
  Knowledge to Robotic Control},'' in \emph{Proc. Conf. Robot Learn. (CoRL)},
  2023, pp. 2165--2183.

\bibitem{rtx2023}
{Open X-Embodiment Collaboration}, ``{Open X-Embodiment: Robotic Learning
  Datasets and RT-X Models},'' arXiv preprint, 2023, arXiv:2310.08864.

\bibitem{moka2024}
K.~Fang, F.~Liu, P.~Abbeel, and S.~Levine, ``{MOKA: Open-World Robotic
  Manipulation through Mark-Based Visual Prompting},'' in \emph{Proc. Robot.:
  Sci. Syst. (RSS)}, 2024.

\bibitem{eureka2023}
Y.~J. Ma \emph{et~al.}, ``{Eureka: Human-Level Reward Design via Coding Large
  Language Models},'' in \emph{Proc. Int. Conf. Learn. Represent. (ICLR)},
  2024.

\bibitem{robocat2023}
K.~Bousmalis \emph{et~al.}, ``{RoboCat: A Self-Improving Generalist Agent for
  Robotic Manipulation},'' \emph{Trans. Mach. Learn. Res.}, 2024.

\bibitem{voyager2023}
G.~Wang \emph{et~al.}, ``{Voyager: An Open-Ended Embodied Agent with Large
  Language Models},'' \emph{Trans. Mach. Learn. Res.}, 2024.

\bibitem{diffusionpolicy2023}
C.~Chi \emph{et~al.}, ``{Diffusion Policy: Visuomotor Policy Learning via
  Action Diffusion},'' in \emph{Proc. Robot.: Sci. Syst. (RSS)}, 2023.

\bibitem{cap2022}
J.~Liang \emph{et~al.}, ``{Code as Policies: Language Model Programs for
  Embodied Control},'' in \emph{Proc. IEEE Int. Conf. Robot. Autom. (ICRA)},
  2023, pp. 9493--9500.

\bibitem{progprompt2022}
I.~Singh \emph{et~al.}, ``{ProgPrompt: Generating Situated Robot Task Plans
  using Large Language Models},'' arXiv preprint, 2022, arXiv:2209.11302.

\bibitem{voxposer2023}
W.~Huang, C.~Wang, R.~Zhang, Y.~Li, J.~Wu, and F.-F. Li, ``{VoxPoser:
  Composable 3D Value Maps for Robotic Manipulation with Language Models},'' in
  \emph{Proc. Conf. Robot Learn. (CoRL)}, 2023, pp. 540--562.

\bibitem{palme2023}
D.~Driess \emph{et~al.}, ``{{P}a{LM}-E: An Embodied Multimodal Language
  Model},'' in \emph{Proc. Int. Conf. Mach. Learn. (ICML)}, 2023, pp.
  8469--8488.

\bibitem{tidybot2023}
J.~Wu \emph{et~al.}, ``{TidyBot: personalized robot assistance with large
  language models},'' \emph{Auton. Robots}, vol.~47, no.~8, pp. 1087--1102,
  2023.

\bibitem{mobilealoha2024}
Z.~Fu, T.~Z. Zhao, and C.~Finn, ``{Mobile ALOHA: Learning Bimanual Mobile
  Manipulation using Low-Cost Whole-Body Teleoperation},'' in \emph{Proc. Conf.
  Robot Learn. (CoRL)}, 2025, pp. 4066--4083.

\bibitem{autort2024}
M.~Ahn \emph{et~al.}, ``{AutoRT: Embodied Foundation Models for Large Scale
  Orchestration of Robotic Agents},'' arXiv preprint, 2024, arXiv:2401.12963.

\bibitem{behavior1k2024}
\BIBentryALTinterwordspacing
C.~Li \emph{et~al.}, ``{BEHAVIOR-1K: A Benchmark for Embodied AI with 1,000
  Everyday Activities and Realistic Simulation},'' in \emph{Proc. Conf. Robot
  Learn. (CoRL)}, ser. Proceedings of Machine Learning Research, vol. 205,
  2023, pp. 80--93. [Online]. Available:
  \url{https://proceedings.mlr.press/v205/li23a.html}
\BIBentrySTDinterwordspacing

\bibitem{ssf2026}
A.~Agarwal \emph{et~al.}, ``{Knowing Where to Look: Interactive Object Search
  via Spatial-Semantic Fusion},'' arXiv preprint, 2026, arXiv:2611.00499.

\end{thebibliography}
\end{document}